\documentclass[10pt,letterpaper,conference]{IEEEtran}

\usepackage{url}
\usepackage{todonotes}
\usepackage{amsmath}
\usepackage{amsfonts}
\usepackage{amssymb}
\usepackage{amsthm}
\usepackage{algpseudocode}
\usepackage{algorithm}
\usepackage{xspace}

\usepackage{tikz}
\usetikzlibrary{plotmarks}
\usepackage[export]{adjustbox}

\usepackage{xcolor,colortbl}
\usepackage{hhline}
\usepackage{wrapfig}
\usepackage{makecell}
\usepackage{multirow}
\usepackage{rotating}

\newcommand{\cla}[1]{\textcolor{black}{#1}}

\newtheorem{definition}{Definition} 
\newtheorem{example}{Example}

\DeclareMathOperator*{\argmin}{argmin}

\newcommand{\R}{\mathbb{R}}

\newcommand{\feats}{\mathcal{X}}
\newcommand{\labels}{\mathcal{Y}}
\newcommand{\hyps}{\mathcal{H}}

\newcommand{\dataset}{\mathcal{D}}
\newcommand{\inst}[3]{(#1,#2)^{#3}}

\newcommand{\xy}{(\vec{x},y)}
\newcommand{\xyk}{\inst{\vec{x}}{y}{k}}
\newcommand{\zy}{(\vec{z},y)}

\newcommand{\Loss}{\ensuremath{\mathcal{L}}\xspace}
\newcommand{\loss}{\ensuremath{\ell}\xspace}

\newcommand{\rewrite}[4]{#1 \xrightarrow{#2}_{#3} #4}

\newcommand{\atkloss}{\Loss^A}

\newcommand{\leaf}[1]{\lambda(#1)}
\newcommand{\node}[1]{\sigma(#1)}

\newcommand{\rootleft}{\dataset_l(f,v,A)}
\newcommand{\rootright}{\dataset_r(f,v,A)}
\newcommand{\rootunk}{\dataset_u(f,v,A)}

\newcommand{\atkunk}{\dataset^\lambda_u(f,v,A)}

\newcommand{\constr}{\mathcal{C}}

\renewcommand\vec{\boldsymbol}

\newcommand{\census}{\texttt{census}}
\newcommand{\wine}{\texttt{wine}}
\newcommand{\credit}{\texttt{credit}}

\newcommand{\stump}{\hat{t}}

\newcommand{\Cl}{\constr_L(\hat{t},A)}
\newcommand{\Cr}{\constr_R(\hat{t},A)}

\newcommand{\binleft}{\dataset^\lambda_L(\hat{t},A)}
\newcommand{\binright}{\dataset^\lambda_R(\hat{t},A)}

\newcommand{\rootbinleft}{\dataset_L(\hat{t},A)}
\newcommand{\rootbinright}{\dataset_R(\hat{t},A)}

\newcommand{\treant}{{\sc Treant}}

\begin{document}

\title{\treant: Training Evasion-Aware Decision Trees}



%
\author{\IEEEauthorblockN{
Stefano Calzavara\IEEEauthorrefmark{1},
Claudio Lucchese\IEEEauthorrefmark{1},
Gabriele Tolomei\IEEEauthorrefmark{2}, 
Seyum Assefa Abebe\IEEEauthorrefmark{1} and
Salvatore Orlando\IEEEauthorrefmark{1}}
\IEEEauthorblockA{\IEEEauthorrefmark{1}Ca' Foscari University of Venice, Italy\\
Email: \{name.surname\}@unive.it}
\IEEEauthorblockA{\IEEEauthorrefmark{2}University of Padua, Italy\\
Email: gtolomei@math.unipd.it}
}

\maketitle

\begin{abstract}
Despite its success and popularity, machine learning is now recognized as vulnerable to \emph{evasion attacks}, i.e., carefully crafted perturbations of test inputs designed to force prediction errors. In this paper we focus on evasion attacks against decision tree ensembles, which are among the most successful predictive models for dealing with non-perceptual problems. Even though they are powerful and interpretable, decision tree ensembles have received only limited attention by the security and machine learning communities so far, leading to a sub-optimal state of the art for adversarial learning techniques. We thus propose \treant, a novel decision tree learning algorithm that, on the basis of a formal threat model, minimizes an evasion-aware loss function at each step of the tree construction. \treant{} is based on two key technical ingredients: \emph{robust splitting} and {\em attack invariance}, which jointly guarantee the soundness of the learning process. Experimental results on three publicly available datasets show that \treant{} is able to generate decision tree ensembles that are at the same time accurate and nearly insensitive to evasion attacks, outperforming state-of-the-art adversarial learning techniques.
\end{abstract}

\section{Introduction}

Machine Learning (ML) is increasingly used in several applications and different contexts. When ML is leveraged to ensure system security, such as in spam filtering and intrusion detection, everybody acknowledges the need of training ML models resilient to adversarial manipulations~\cite{HuangJNRT11,BiggioR17}. Yet the same applies to other critical application scenarios in which ML is now employed, where adversaries may cause severe system malfunctioning or faults. For example, consider an ML model which is used by a bank to grant loans to inquiring customers: a malicious user may try to fool the model into illicitly qualifying him for a loan. Unfortunately, traditional ML algorithms proved vulnerable to a wide range of attacks, and in particular to \emph{evasion attacks}, i.e., carefully crafted perturbations of test inputs designed to force prediction errors~\cite{BiggioCMNSLGR13,NguyenYC15,PapernotMJFCS16,Moosavi-Dezfooli16}. 

To date, research on evasion attacks has mostly focused on linear classifiers~\cite{LowdM05,BiggioNL11} and, more recently, on deep neural networks~\cite{SzegedyZSBEGF13,GoodfellowSS15}. Whereas deep learning obtained remarkable and revolutionary results on many perceptual problems, such as those related to computer vision and natural language understanding, \emph{decision trees ensembles} are nowadays one of the best methods for dealing with non-perceptual problems, and are one of the most commonly used techniques in
Kaggle competitions~\cite{Chollet:2017:DLP:3203489}. Decision trees are also \emph{interpretable} models~\cite{tolomei2017kdd}, yielding predictions which are human-understandable in terms of syntactic checks over domain features, which is particularly appealing in the security setting. Unfortunately, despite their success, decision tree ensembles have received only limited attention by the security and machine learning communities so far, leading to a sub-optimal state of the art for adversarial learning techniques (see Section~\ref{sec:related}).

In this paper, we propose \treant,\footnote{The name comes from the role playing game ``Dungeons \& Dragons'', where it identifies giant tree-like creatures.} a novel learning algorithm designed to build decision trees which are resilient against evasion attacks at test time. Based on a formal threat model, \treant{} optimizes an evasion-aware loss function at each step of the tree construction~\cite{MadryMSTV17}. This is particularly challenging to enforce correctly, considered the greedy nature of traditional decision tree learning~\cite{hunt1966experiments}. In particular, \treant{} has to ensure that the local greedy choices performed upon tree construction are not short-sighted with respect to the capabilities of the attacker, who has the advantage of choosing the best attack strategy based on the fully built tree. \treant{} is based on the combination of two key technical ingredients: a \emph{robust splitting} strategy for decision tree nodes, which reliably takes into account at training time the attacker's capability of perturbing instances at test time, and an \emph{attack invariance} property, which preserves the correctness of the greedy construction by generating and propagating constraints along the decision tree, so as to discard splitting choices which might be vulnerable to attacks. 

We finally deploy our learning algorithm within a traditional random forest framework~\cite{RandomF01} and show its predictive power on real-world datasets. Notice that, although there have been various proposals that tried to improve robustness against evasion attacks by using ensemble methods~\cite{Multiple05,Multiple06,Multiple08,BiggioFR10}, it was shown that ensembles of weak models are not necessarily strong~\cite{HeWCCS17}. We avoid this shortcoming by employing \treant{} to train an ensemble of decision trees which are individually resilient to evasion attempts.

\subsection{Roadmap}
To show how \treant{} improves over the state of the art, we proceed as follows:
\begin{enumerate}
\item We first review decision trees and decision tree ensembles, presenting a thorough critique of existing adversarial learning techniques for such models (Section~\ref{sec:background}).
\item We introduce our formal threat model, discussing an exhaustive white-box attack generation method, which allows for an accurate evaluation of the performance of decision trees under attack and proves scalable enough for our experimental analysis (Section~\ref{sec:threat}).
\item We present \treant, the first tree learning algorithm which greedily, yet soundly, minimizes an evasion-aware loss function upon tree construction (Section~\ref{sec:algorithm}).
\item We experimentally show that \treant{} outperforms existing adversarial learning techniques on three publicly available datasets (Section~\ref{sec:experiments}). 
\end{enumerate}

Our analysis shows that \treant{} is able to build decision tree ensembles that are at the same time accurate and nearly insensitive to evasion attacks. Compared to the state of the art,
\treant{} exhibits a ROC AUC improvement against the strongest attacker ranging from $\approx 10\%$ to $\approx 20\%$.

\section{Background and Related Work}
\label{sec:background}

\subsection{Supervised Learning}
Let $\feats \subseteq \R^d$ be a $d$-dimensional vector space of real-valued \textit{features}. An {\em instance} $\vec{x} \in \feats$ is a $d$-dimensional feature vector $(x_1, x_2, \ldots, x_d)$ representing an object in the vector space.\footnote{For simplicity, we only consider numerical features over $\R$. However, our framework can be readily generalized to other use cases, e.g., categorical or ordinal features, which we support in our implementation and experiments.} Each instance $\vec{x} \in \feats$ is assigned a label $y \in \labels$ by some unknown function $g: \feats \mapsto \labels$, called the \emph{target} function. Starting from a set of hypotheses $\hyps$, the goal of a \emph{supervised learning} algorithm is to find the function $\hat{h} \in \hyps$ that best approximates the target $g$.
This is practically achieved through empirical risk minimization~\cite{vapnik1992principles}; given a sample of correctly labeled instances $\dataset =$ $ \{(\vec{x}_1, g(\vec{x}_1)), \ldots, (\vec{x}_n, g(\vec{x}_n))\}$ known as the \emph{training set}, the empirical risk is defined by a loss function $\Loss: \hyps \times (\feats \times \labels)^n \mapsto \R^+$ measuring the cost of erroneous predictions, i.e., the cost of predicting $\hat{h}(\vec{x}_i)$ instead of the true label $g(\vec{x}_i)$, for all $(\vec{x}_i, g(\vec{x}_i)) \in \dataset$. Supervised learning thus amounts to finding:
\[
\hat{h} = \argmin_{h \in \hyps} \Loss(h, \dataset). 
\]

The loss \Loss is often obtained by aggregating an instance-level loss $\loss: \labels \times \labels \mapsto \R^+$. Here, we define $\Loss$ as the sum of the instance-level losses: $\Loss(h, \dataset) = \sum_{(\vec{x},y) \in \dataset} \loss(h(\vec{x}), y)$.

\subsection{Decision Trees and Decision Tree Ensembles}
A powerful set of hypotheses $\hyps$ is the set of the \emph{decision trees}~\cite{BreimanFOS84}. We focus on binary decision trees, whose internal nodes perform thresholding over feature values. Such trees can be inductively defined as follows: a decision tree $t$ is either a leaf $\leaf{\hat{y}}$ for some label $\hat{y} \in \labels$ or a non-leaf node $\node{f,v,t_l,t_r}$, where $f \in [1,d]$ identifies a feature, $v \in \R$ is the threshold for the feature $f$ and $t_l,t_r$ are decision trees. At test time, an instance $\vec{x}$ traverses the tree $t$ until it reaches a leaf $\leaf{\hat{y}}$, which returns the \textit{prediction} $\hat{y}$, denoted by $t(\vec{x}) =  \hat{y}$. Specifically, for each traversed tree node $\node{f,v,t_l,t_r}$, $\vec{x}$ falls into the left tree $t_l$ if $x_f \leq v$, and into the right tree $t_r$ otherwise. We just write $\lambda$ or $\sigma$ to refer to some leaf or node of the decision tree when its actual content is irrelevant. The problem of learning an optimal decision tree is known to be NP-complete~\cite{HyafilR76,Murthy98}; as such, a top-down greedy approach is usually adopted~\cite{hunt1966experiments}, as shown in Algorithm~\ref{alg:basictree}. 

\begin{algorithm}[t]
\caption{\textsc{BuildTree}}
\label{alg:basictree}
\begin{algorithmic}[1]
\State {\bfseries Input:} {training data $\dataset$}

\State $\hat{y} \gets \argmin_y \Loss(\leaf{y},\dataset)$

\State $\node{f,v,\leaf{\hat{y}_l},\leaf{\hat{y}_r}},\dataset_l,\dataset_r \gets$ {\sc BestSplit}$(\dataset)$

\If{$\Loss(\node{f,v,\leaf{\hat{y}_l},\leaf{\hat{y}_r}},\dataset) < \Loss(\leaf{\hat{y}},\dataset)$}
	\State {$t_l \gets$ {\sc BuildTree}$({\dataset_l})$}
	\State {$t_r \gets$ {\sc BuildTree}$({\dataset_r})$}
    \State {\bf return} $\node{f,v,t_l,t_r}$
\Else
    \State {\bf return} $\leaf{\hat{y}}$
\EndIf
\end{algorithmic}
\end{algorithm}

\begin{algorithm*}[t]
\caption{\textsc{BestSplit}} 
\label{alg:bestsplit}
\begin{algorithmic}[1]
\State {\bfseries Input:} training data $\dataset$

\Comment{Build a set of candidate tree nodes $\mathcal{N}$ via an exhaustive search over $f$ and $v$}
\State{$\mathcal{N} \gets \{ \sigma(f,v,\leaf{\hat{y}_l}, \leaf{\hat{y}_r}) ~|~ f \in [1,d] \wedge \exists \xy \in \dataset: x_f = v ~\wedge \hat{y}_l,\hat{y}_r = \argmin_{y_l, y_r} \Loss(\sigma(f,v,\leaf{y_l}, \leaf{y_r}), \dataset)\}$}\label{alg:search}

\Comment{Select the candidate node $\hat{t} \in \mathcal{N}$ which minimizes the loss $\Loss$ on the training data $\dataset$}
\State{\label{alg:sel}$\hat{t} = \argmin_{t \in \mathcal{N}} \Loss(t, \dataset) = \sigma(f,v,\leaf{\hat{y}_l},\leaf{\hat{y}_r})$
}

\Comment{Split the training data $\dataset$ based on the best candidate node $\hat{t} = \sigma(f,v,\leaf{\hat{y}_l},\leaf{\hat{y}_r})$}
\State{$\dataset_l \gets \{(\vec{x},y) \in \dataset ~|~ x_f \leq v\}$ }\label{alg:d-part}

\State $\dataset_r \gets  \dataset \setminus \dataset_l$

\State {\bf return} {$\hat{t},\dataset_l,\dataset_r$}

\end{algorithmic}
\end{algorithm*}

The function \textsc{BuildTree} takes as input a dataset $\dataset$ and initially computes the label $\hat{y}$ which minimizes the loss on $\dataset$ for a decision tree composed of just a single leaf; for instance, when the loss is the Sum of Squared Errors (SSE), such label just amounts to the mean of the labels in $\dataset$. The function then checks if it is possible to grow the tree to further reduce the loss by calling a \emph{splitting} function \textsc{BestSplit} (Algorithm~\ref{alg:bestsplit}), which attempts to replace the leaf $\leaf{\hat{y}}$ with a new sub-tree $\node{f,v,\leaf{\hat{y}_l},\leaf{\hat{y}_r}}$. This sub-tree is greedily identified by choosing $f$ and $v$ from an exhaustive exploration of the search space consisting of all the possible features and thresholds, and with the predictions $\hat{y}_l$ and $\hat{y}_r$ chosen
so as to minimize the global loss on $\dataset$.
If it is possible to reduce the loss on $\dataset$ by growing the new sub-tree, the tree construction is recursively performed over the subsets $\dataset_l = \{(\vec{x},y) \in \dataset ~|~ x_f \leq v\}$ and $\dataset_r = \dataset \setminus \dataset_l$, otherwise the original leaf $\leaf{\hat{y}}$ is returned. 
Real-world implementations of the algorithm typically use multiple stopping criteria to prevent overfitting, e.g., by bounding the tree depth, or by requiring a minimum number of instances in the datasets used in the recursive calls.

Random Forest (RF) and Gradient Boosting Decision Trees (GBDT) are popular ensemble learning methods for decision trees~\cite{RandomF01,friedman2001greedy}. RFs are obtained by independently training a set of trees $\mathcal{T}$, which are combined into the \emph{ensemble predictor} $\hat{h}$, e.g., by using majority voting to assign the class label. Each $t_i \in \mathcal{T}$ is typically built by using bagging and per-node feature sampling over the training set. In GBDTs, instead, each tree approximates a gradient descent step along the direction of loss minimization. Both methods are very effective, where RF is able to train models with low variance, while GDBTs are models of high accuracy yet possibly prone to overfit.

\subsection{Related Work}
\label{sec:related}
Adversarial learning, which investigates the safe adoption of ML in adversarial settings~\cite{HuangJNRT11}, is a research field that has been consistently increasing of importance in the last few years. In this paper we deal with \emph{evasion attacks}, a research sub-field of adversarial learning, where deployed ML models are targeted by attackers who craft adversarial examples that resemble normal data instances, but force wrong predictions. Most of the work in this field regards classifiers, in particular binary ones. The attacker starts from a positive instance that is classified correctly by the deployed ML model and is interested in introducing minimal perturbations on the instance to modify the prediction from positive to negative, thus ``evading'' the classifier~\cite{NelsonRHJLLRTT10,BiggioCMNSLGR13,BiggioFR14,SrndicL14,KantchelianTJ16,Carlini017,DangHC17,GoodfellowSS15}. 

To prevent these attacks, different techniques have been proposed for different models, including support vector machines~\cite{BiggioNL11,XiaoBNXER15}, deep neural networks~\cite{GuR14,GoodfellowSS15,PapernotM0JS16}, and decision tree ensembles~\cite{KantchelianTJ16,ChenZBH19}. Unfortunately, the state of the art for decision tree ensembles is far from satisfactory. 

The first adversarial learning technique for decision tree ensembles is due to Kantchelian {\em et al.} and is called \emph{adversarial boosting}~\cite{KantchelianTJ16}. It is an empirical data augmentation technique,
borrowing from the {\em adversarial training} approach \cite{SzegedyZSBEGF13},
where a number of evading instances are included among the training data to make the learned model aware of the attacks and, thereby, possibly more resilient to them. Specifically, at each boosting round, the training set is extended by crafting a set of possible perturbations for each original instance and by picking the one with the smallest margin, i.e., the largest misprediction risk, for the model trained that far. Adding perturbed instances to the training set forces the learning algorithm to minimize the {\em average} error over both the original instances and the chosen sample of evading ones, but this does not provide clear performance guarantees under attack. This is both because evading instances exploited at training time might not be representative of test-time attacks, and because optimizing the average case might not defend against the {\em worst-case} attack. Indeed, the experiments in Section~\ref{sec:experiments} show that the performance of ensembles trained via adversarial boosting can be severely downgraded by evasion attacks.

The second adversarial learning technique for decision tree ensembles was proposed in a very recent work by Chen {\em et al.}, who introduced the first tree learning algorithm embedding the attacker directly in the optimization problem solved upon tree construction~\cite{ChenZBH19}. 
The key idea of their approach, called \emph{robust trees}, is to redefine the splitting strategy of the training examples at a tree node. They first identify the so called {\em unknown} instances of $\dataset$, which may fall in either in $\dataset_l$ or in $\dataset_r$, depending on adversarial perturbations. 
The authors thus claim that the optimal tree construction strategy would need to account for an exponential number of attack configurations over these unknown instances. 
To tame such algorithmic complexity, they propose a sub-optimal heuristic approach based on four ``representative'' attack cases. Though the key idea of this algorithm is certainly interesting and shares some similarities with our own proposal, it also suffers from significant shortcomings. First, representative attack cases are not such anymore when the attacker is aware of the defense mechanism,
and they are not anyway sufficient to subsume the spectrum of possible attacks: our algorithm takes into account all the possible attack cases, while being efficient enough for practical adoption. Moreover, the approach in~\cite{ChenZBH19} does not implement safeguards against the incremental greedy nature of decision tree learning: there is no guarantee that, once the best splitting has been identified, the attacker cannot adapt his strategy to achieve better results on the full tree. Indeed, the experimental evaluation in Section~\ref{sec:experiments} shows that it is very easy to evade the trained models, which turn out to be even more fragile than those trained through adversarial boosting~\cite{KantchelianTJ16}.

\section{Threat Model}
\label{sec:threat}

The possibility to craft adversarial examples was popularized by Szegedy {\em et al.} in the image classification domain: their seminal work showed that it is possible to introduce minimal perturbations into an image so as to modify the prediction of its class by a deep neural network~\cite{SzegedyZSBEGF13}.
These evasion attacks questioned the applicability of ML to several security/business critical domains where malicious users can intentionally fool an ML model deployed online.

\subsection{Loss Under Attack and Adversarial Learning}

At an abstract level, we can see the attacker $A$ as a function mapping each instance to a set of possible perturbations, which might be able to evade the ML model. Depending on the specific application scenario, not every attack is plausible, e.g., $A$ cannot force some perturbations or behaves surreptitiously to avoid detection. For instance, in the typical image classification scenario, $A$ is usually assumed to introduce just slight modifications that are perceptually undetectable to humans. This simple similarity constraint between the original instance $\vec{x}$ and its perturbed variant $\vec{z}$ is well captured by a distance~\cite{GoodfellowSS15}, i.e., we might have $A(\vec{x}) =\{\vec{z} ~|~\|\vec{z}-\vec{x}\|_\infty \leq \epsilon\}$.

Similarly, assuming that the attacker can run independent attacks on every instance of a given dataset $\dataset$, we can define $A(\dataset)$ as the set of the datasets $\dataset'$ which can be obtained by replacing each $\xy \in \dataset$ with any $\zy$ such that $\vec{z} \in A(\vec{x})$.

The easiness of crafting successful evasion attacks defines the {\em robustness} of a given ML model at test time. The goal of learning a robust model is therefore to minimize the harm an attacker may cause via perturbations. This learning goal was formalized as a min-max problem by Madry {\em et al.}~\cite{MadryMSTV17}:
\begin{equation}
\label{eq:minmax}
\hat{h} = \argmin_{h \in \hyps}\quad  \underbrace{\max_{\dataset' \in A(\dataset)} \Loss(h,\dataset')}_{\atkloss(h,\dataset)}.
\end{equation}

The inner maximization problem models the attacker $A$ replacing all the given instances with an adversarial example aimed at maximizing the loss. We call {\em loss under attack}, noted $\atkloss(h,\dataset)$, the solution to the inner maximization problem. The outer minimization resorts to the empirical risk minimization principle, aiming to find the hypothesis that minimizes the loss under attack on the training set. 

\subsection{Attacker Model}
Distance-based constraints for defining the attacker's capabilities are very flexible for perceptual problems and proved amenable for heuristic algorithms for solving the inner maximization problem of Equation~\ref{eq:minmax}~\cite{MadryMSTV17}. However, they cannot be easily generalized to other realistic application scenarios, e.g., where perturbations are not symmetric, where the attacker may not be able to alter some of the features, or where categorical attributes are present. To overcome such limitations, 
we model the attacker $A$ as a pair $(R, K)$, where $R$ is a set of \emph{rewriting rules}, defining how instances can be corrupted, and $K \in \R^+$ is a \emph{budget}, limiting the amount of alteration the attacker can apply to individual instances. Each rule $r \in R$ has the form:
\begin{equation*}
\rewrite{[a,b]}{f}{k}{[\delta_l,\delta_u]},
\end{equation*}
where $[a,b]$ and $[\delta_l,\delta_u]$ are intervals on $\R \cup \{-\infty,+\infty\}$, with the former defining the \emph{precondition} for the application of the rule and the latter defining the \emph{magnitude} of the perturbation enabled by the rule; $f \in [1,d]$ is the index of the feature to corrupt; and $k \in \R^+$ is the \emph{cost} of the rule. The semantics of the rewriting rule can be explained as follows: if an instance $\vec{x}$ satisfies the condition $x_f \in [a,b]$, then the attacker can corrupt the instance $\vec{x}$ by adding any $v \in [\delta_l,\delta_u]$ to $x_f$ and spending $k$ from the available budget. The attacker can corrupt each instance by using as many rewriting rules as desired in whatever order, up to budget exhaustion. 


According to this attacker model, we define $A(\vec{x})$, the set of the attacks against an instance $\vec{x}$, as follows.

\begin{definition}[Attacks]
Given an instance $\vec{x}$ and an attacker $A = (R,K)$, we let $A(\vec{x})$ be the set of the \emph{attacks} that can be obtained from $\vec{x}$, i.e., the set of the instances $\vec{z}$ such that there exists a sequence of rewriting rules $r_1,\ldots,r_n \in R$ and a sequence of instances $\vec{x}_0,\ldots,\vec{x}_n$ where:
\begin{enumerate} 
\item $\vec{x}_0 = \vec{x}$ and $\vec{x}_n = \vec{z}$;
\item for all $i \in [1,n]$, the instance $\vec{x}_{i-1}$ can be corrupted into the instance $\vec{x}_i$ by using the rewriting rule $r_i$;
\item the sum of the costs of $r_1,\ldots,r_n$ is not greater than $K$.
\end{enumerate} 
Notice that $\vec{x} \in A(\vec{x})$ for any attacker $A$ by picking an empty sequence of rewriting rules.
\end{definition}

 We highlight that this rule-based attacker model includes novel attack capabilities like asymmetric perturbations, easily generalizes to categorical variables, and still covers or approximates standard distanced-based models. For instance, $L^0$-norm attacker models where the attacker can corrupt at will a limited number of features can be easily represented~\cite{KantchelianTJ16}. The use of a budget is convenient to fine-tune the power of the attacker and enables the adoption of standard evaluation techniques for ML models under attack, like \emph{security evaluation curves}~\cite{BiggioR17}.

\subsection{Attack Generation}
\label{sec:attacks}
Computing the loss under attack $\atkloss$ is useful to evaluate the resilience of ML models to evasion attacks at test time; yet this might be intractable, since it assumes the ability to identify the most effective attack for all the test instances. This issue is thus typically dealt with by using a heuristic attack generation algorithm, e.g., the fast gradient sign method~\cite{GoodfellowSS15} or any of its variants, to craft adversarial examples which empirically work well. However, our focus on decision trees and the adoption of a rule-based attacker model enables an exhaustive attack generation strategy for the test set which, though computationally expensive, proves scalable enough for our experimental analysis and allows the actual identification of the most effective attacks. This enables the most accurate security assessment in terms of the actual value of $\atkloss$.

We consider a {\em white-box} attacker model, where the attacker has a complete knowledge of the trained decision tree ensemble. We thus assume that the attacker exploits the knowledge of the structure of the trees in the targeted ensemble and, most importantly, of the features and thresholds which are actually used in the prediction process.
Note that a decision tree ensemble induces a finite partitioning of the input vector space $\feats$,
defined by the features and thresholds used in the internal nodes of the trees in the ensemble,
where instances falling in the same partition share the same prediction.
This partitioning makes it possible to significantly reduce the set of attacks that are relevant to the computation of $\atkloss$ by considering at most one representative attacked instance for a given partition.
We achieve this by a recursive algorithm that, for the sake of space, we just sketch below.
For any given instance $\vec{x}$, we recursively apply all valid rules up to budget exhaustion.
In doing so, the interval $[x_f+\delta_l,x_f+\delta_u]$ of each applied rewriting rule, is split into sub-intervals induced by the ensemble's thresholds relative to feature $f$, and we generate a single attack for each of the sub-intervals,
including the extremes $x_f+\delta_l,x_f+\delta_u$. Note that we include the extremes of the preconditions of the rewriting rules in the partitioning, as to make sure that all recursively applicable rules are considered. The above enumeration strategy makes sure that all relevant attacks, i.e, causing at least one internal node of the ensemble to invert its outcome, are generated.



\section{\treant: Learning Robust Decision Trees}
\label{sec:algorithm}

\begin{figure*}[tb] 
\begin{center}

\adjincludegraphics[height=14cm,trim={0 0 0 0},clip]{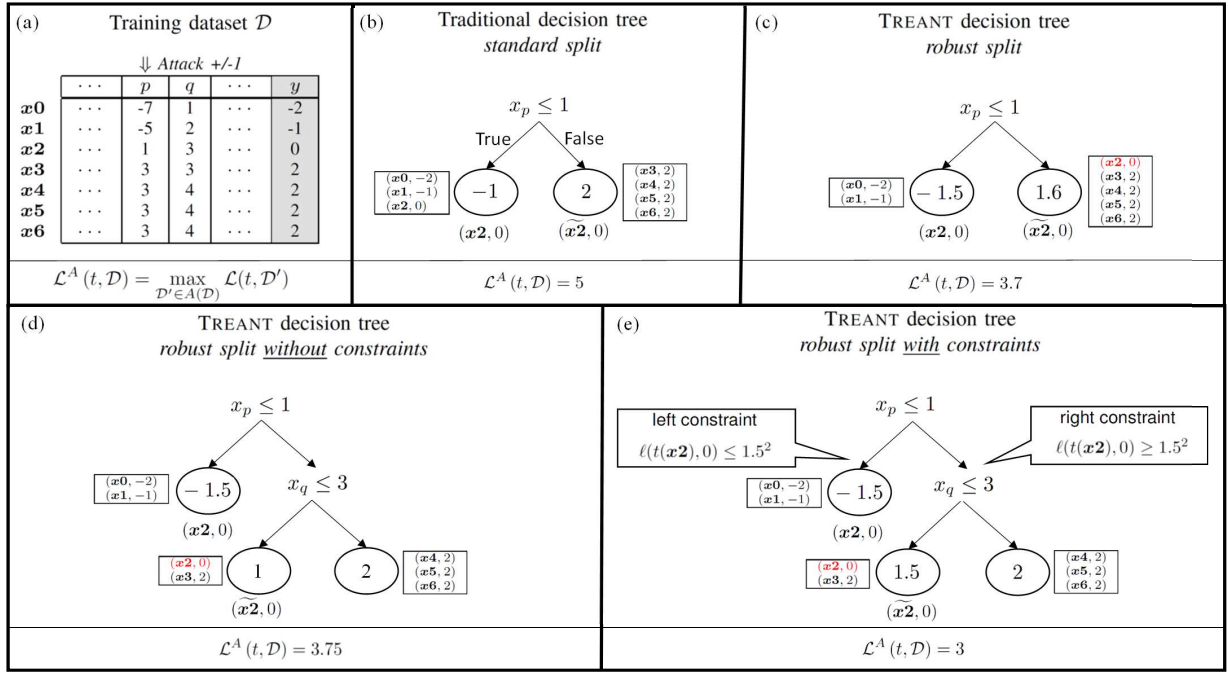}\vspace{-2.7cm}
\end{center}
\caption{Overview of the \treant{} construction and its key challenges.}
\label{fig:ex}
\end{figure*}

In this section, we present a novel decision tree learning algorithm that, by minimizing the loss under attack $\atkloss$ at training time, enforces resilience to evasion attacks at test time. We call \treant{} the proposed algorithm.

\subsection{Overview}
Compared to Algorithm~\ref{alg:basictree}, \treant{} replaces the {\sc BestSplit} function by revising $(i)$ the computation of predictions on the new leaves, $(ii)$ the selection of the best split and $(iii)$ the dataset partition along the recursion.

Before discussing the technical details, we build on the toy example in Figure~\ref{fig:ex} to illustrate the non-trivial issues arising when optimizing $\atkloss$. Figure~\ref{fig:ex}.(a) shows a dataset $\dataset$ for which we assume the attacker $A=(\{r\},1)$, where $r$ is a rewriting rule of cost 1 which allows the corruption of the feature $p$ by adding any value in the interval $[-1,+1]$.

Assuming SSE is used as the underlying loss function \Loss, the decision stump initially generated by Algorithm~\ref{alg:basictree} is shown in Figure~\ref{fig:ex}.(b) along with the result of the splitting. Note that while the loss $\Loss = 2$ is small,\footnote{$\Loss(t,\dataset) =  (-2+1)^2+(-1+1)^2+(-1-0)^2+4\cdot(2-2)^2=2$.}
the loss under attack $\atkloss = 5$ is much larger.\footnote{$\atkloss(t, \dataset) =  (-2+1)^2+(-1+1)^2+(2-0)^2+4\cdot(2-2)^2=5$.} 
This is because the attacker may alter $\vec{x2}$ into a perturbed instance $\widetilde{\vec{x2}}$ so as to reverse the outcome of the test $x_p \leq 1$, i.e., the original instance $\vec{x2}$ falls into the left leaf of the stump, but the perturbed instance $\widetilde{\vec{x2}}$ falls into the right leaf. The first issue of Algorithm~\ref{alg:basictree} is thus that the estimated loss $\Loss$ on the training set, computed when building the decision stump, is smaller than the loss under attack $\atkloss$ we would like to minimize. We solve this issue by designing a novel \emph{robust splitting} strategy to identify the best split of $\dataset$, which directly minimizes $\atkloss$ when computing the leaves predictions and leads to the generation of a tree that is more robust to attacks. In particular, the decision stump learnt by using our robust splitting strategy is shown in Figure~\ref{fig:ex}.(c), where the leaves predictions have been found by assuming that $\vec{x2}$ actually falls into the right leaf (according to the best attack strategy). For this new decision stump, the best move for the attacker is still to corrupt $\vec{x2}$, but the resulting $\atkloss = 3.7$ is much smaller than that of the previous stump.\footnote{$\atkloss(t,\dataset) = (-2+1.5)^2+(-1+1.5)^2+(0-1.6)^2+4\cdot(2-1.6)^2=3.7$.} The figure also shows the outcome of the robust splitting.

However, a second significant issue arises when the decision stump is recursively grown into a full decision tree. Suppose to further split the right leaf of Figure~\ref{fig:ex}.(c), therefore considering only the instances falling therein, including the instance $\vec{x2}$ put there by the robust splitting. We would find that the best split is given by $x_q \le 3$, where the feature $q$ cannot be modified by the attacker. The resulting tree is shown in Figure~\ref{fig:ex}.(d). Note however that, by creating the new sub-tree, new attacking opportunities show up, because the attacker now finds more convenient to just leave $\vec{x2}$ unaltered and let it fall directly into the left child of the root. As a consequence, by adding the new sub-tree, we observe an increased loss under attack $\atkloss = 3.75$.\footnote{$\Loss^A(t,\dataset) = (-2+1.5)^2+(-1+1.5)^2+(0-1.5)^2+(2-1)^2+3\cdot(2-2)^2=3.75$.} This second issue can be solved by ensuring that any new sub-tree does not create new attacking opportunities that generate a larger loss. We call this property {\em attack invariance}. The proposed algorithm grows the sub-tree on the right leaf by carefully adjusting its predictions as shown in Figure~\ref{fig:ex}.(e), still decreasing the loss under attack to $\atkloss = 3$ with respect to the tree in Figure~\ref{fig:ex}.(c).\footnote{$\Loss^A(t,\dataset) = (-2+1.5)^2+(-1+1.5)^2+(0-1.5)^2+(2-1.5)^2+3\cdot(2-2)^2=3$.} This is enforced by including constraints along the tree construction, as shown in the figure.


To sum up, the key technical ingredients of \treant{} are:
\begin{enumerate}
\item \emph{Robust splitting}: given a candidate feature $f$ and threshold $v$, the robust splitting strategy evaluates the quality of the corresponding node split on the basis of a \emph{ternary} partitioning of the instances falling into the node. It identifies those instances for which the outcome of the node predicate $x_f \le v$ depends on the attacker's moves, and those that cannot be affected by the attacker, thus always traversing the left or the right branch of the new node. In particular, the $\atkloss$ minimization problem is reformulated on the basis of left, right and unknown instances, i.e., instances which might fall either left or right depending on the attacker. Finally, the recursion on the left and right child of the node is performed by separating the instances in a binary partition based on the effects of the most harmful attack (Section~\ref{sec:robsplit}).

\item \emph{Attack invariance}: a security property requiring that the addition of a new sub-tree does not allow the attacker to find better attack strategies that increase $\atkloss$. Attack invariance is achieved by imposing an appropriate set of constraints upon node splitting. New constraints are generated for each of the attacked instances present in the split node and are propagated to the child nodes upon recursion (Section~\ref{sec:atk-invariance}).
\end{enumerate}

The pseudo-code of the algorithm is given in Section~\ref{sec:pseudocode}. To assist the reader, the notation used in the present section is summarized in Table~\ref{tab:notation}.

\begin{table}[tb]
    \centering
    \caption{\label{tab:notation} Notation Summary}
    \begin{tabular}{@{}c|p{.6\columnwidth}@{}}
        Symbol & Meaning \\ \hline
        $\dataset$ & Training dataset \\
        $\dataset^\lambda$ & Local projection of $\dataset$ on the leaf $\lambda$ \\
        $A(\vec{x})$ & Set of all the attacks $A$ can generate from $\vec{x}$\\ 
        $A(\dataset)$ & Set of all the attacks $A$ can generate from $\dataset$ \\
        $\leaf{\hat{y}}$ & Leaf node with prediction $\hat{y}$ \\
        $\node{f,v,t_l,t_r}$ & Node testing $x_f\!\le\!v$, having sub-trees $t_l,t_r$ \\
        $\rootleft$ & Left elems of ternary partitioning on $(f,v)$ \\
        $\rootright$ & Right elems of ternary partitioning on $(f,v)$ \\
        $\rootunk$ & Unkn. elems of ternary partitioning on $(f,v)$ \\
        $\rootbinleft$ & Left elems of robust splitting on $\hat{t}$ \\
        $\rootbinright$ & Right elems of robust splitting on $\hat{t}$ \\
        $\Cl$ & Set of constraints for the left child of $\hat{t}$ \\
        $\Cr$ & Set of constraints for the right child of $\hat{t}$ \\
    \end{tabular}
\end{table}

\subsection{Robust Splitting}
\label{sec:robsplit}
We present our novel {\em robust splitting} strategy that grows the current tree $t$ by replacing a leaf $\lambda$ with a new sub-tree so as to minimize the loss under attack $\atkloss$. For the sake of clarity, we discuss it as if the splitting was employed on the root node of a new tree, i.e., to learn the decision stump that provides the best loss reduction on the full input dataset $\dataset$. The next subsection discusses the application of the proposed strategy during the recursive steps of the tree-growing process.

Aiming at greedily optimizing the  \textit{min-max} problem in Equation~\ref{eq:minmax}, we have to find the best decision stump $\stump = \node{f,v,\leaf{\hat{y}_l},\leaf{\hat{y}_r}}$ such that:
\[
\begin{array}{lcl}
 \stump &=& \argmin\limits_{t}  \atkloss\left(t, \dataset\right) = \\
 & = & \argmin\limits_{t} \max\limits_{\dataset' \in A(\dataset)} \Loss(t,\dataset') = \\[.3cm]
& = &  \argmin\limits_{t} \displaystyle\sum\limits_{\xy \in \dataset} \ \ \ \max_{\vec{z} \in A(\vec{x})} \loss(t(\vec{z}), y).
\end{array}
\]

Whereas the pair $(f,v)$ in $\stump = \node{f,v,\leaf{\hat{y}_l},\leaf{\hat{y}_r}}$ can be determined via an exhaustive search, the predictions $\hat{y}_l$ and $\hat{y}_r$ must be found by minimizing the loss under attack $\atkloss$. However, this is not trivial, because the loss incurred by an instance $\xy$ may depend on the attacks it is possibly subject to.
Similarly to~\cite{ChenZBH19}, we thus define a ternary partitioning of the training dataset as follows.

\begin{definition}[Ternary Partitioning]
For a feature $f$, a threshold $v$ and an attacker $A$, the \emph{ternary partitioning} of the dataset $\dataset = \rootleft \cup \rootright \cup \rootunk$ is defined by:
\[
\begin{array}{l}
\rootleft  =  \{\xy \in \dataset ~|~ \forall \vec{z} \in A(\vec{x}): z_f \leq v\} \\
\rootright = \{\xy \in \dataset ~|~ \forall \vec{z} \in A(\vec{x}): z_f > v\} \\
\rootunk   = (\dataset \setminus \rootleft) \setminus \rootright.
\end{array}
\]
\end{definition}

In words, $\rootleft$ includes those instances $\xy$ falling into the left branch regardless of the attack, hence the attacker has no gain in perturbing $x_f$. A symmetric reasoning applies to $\rootright$, containing those instances which fall into the right branch for all the possible attacks. The instances that the attacker may actually want to target are those falling into $\rootunk$, thus aiming at the largest loss. By altering those instances, the attacker may force each $\xy \in \rootunk$ to fall into the left
branch with a loss of $\loss(\hat{y}_l,y)$, 
or into the right branch, with a loss of $\loss(\hat{y}_r,y)$.

\begin{example}[Ternary Partitioning]
The test node $x_p \le 1$ and the attacker considered in Figure~\ref{fig:ex}.(c) determine the following ternary partitioning of $\dataset$:
\begin{itemize}
    \item $\dataset_l(p,1,A) = \{(\vec{x0},-2), (\vec{x1},-1)\}$
    \item $\dataset_r(p,1,A) = \{(\vec{x3},2), (\vec{x4},2), (\vec{x5},2), (\vec{x6},2)\}$
    \item $\dataset_u(p,1,A) = \{(\vec{x2},0)\}$
\end{itemize}

In other words, the instance $\vec{x2}$ is the only instance for which the branch taken at test time is unknown, as it depends on the attacker $A$. 
\end{example}

By construction, given $(f,v)$, the loss $\atkloss$ can be affected by the presence of the attacker $A$ only for the instances in $\rootunk$, while for all the remaining instances it holds that $\atkloss = \Loss$. Since the attacker may force each instance of $\rootunk$ to fall into either the left or the right branch, the authors of~\cite{ChenZBH19} acknowledge a combinatorial explosion in the computation of $\atkloss$. Rather than evaluating all the possible configurations, they thus propose a heuristic approach evaluating four ``representative'' attack cases: {\em i)} no attack, {\em ii)} all the unknown instances are forced in the left child, {\em iii)} all the unknown instances are forced in the right child, and {\em iv)} all the unknown instances are swapped by the attacker, i.e., they are forced in the left/right child when they would normally fall in the right/left child. Then, the loss $\Loss$ is evaluated for these four split configurations and the maximum is used to estimate $\atkloss$, so as to find the best stump $\hat{t}$ to grow. Note that $\Loss$ is computed as in a standard decision tree learning algorithm. Unfortunately, this heuristic strategy does not offer soundness guarantees, because the above four configurations leave potentially harmful attacks out of sight and do not induce an upper-bound of $\atkloss$. 

To avoid this soundness issue, while keeping the tree construction tractable, we  pursue a \emph{numerical optimization} as follows. For a given $(f,v)$, we highlight that finding the best attack configuration and finding the best left/right leaves predictions ${\hat{y}}_l,{\hat{y}}_r$ are two inter-dependent problems, yet the strategy adopted in~\cite{ChenZBH19} is to first evaluate a few different attack configurations, and then to find the leaves predictions. We instead solve these two problems simultaneously
by using a formulation of the min-max problem that, fixed $(f,v)$,
is expressed solely in terms of ${\hat{y}}_l,{\hat{y}}_r$:
 	\begin{equation}
 	\label{eq:optimization}
 	({\hat{y}}_l,{\hat{y}}_r) = \argmin\limits_{y_l,y_r}  \atkloss(\node{f,v,\leaf{y_l},\leaf{y_r}},\dataset),
 	\end{equation}
 	where $\atkloss$ is decomposed via the ternary partitioning as:
 	\[
 	\begin{array}{l}
 	\atkloss(\node{f,v,\leaf{y_l},\leaf{y_r}},\dataset) \quad = \\[.2cm]
 	\quad \quad = \quad \Loss(\leaf{{y}_l},\rootleft) +  \Loss(\leaf{{y}_r},\rootright)\ + \\[.2cm]
 	\quad \quad \quad \quad + \displaystyle\sum\limits_{\xy \in \rootunk} \max \{\ell({y}_l,y), \ell({y}_r,y)\}.
 	\end{array}
 	\]

Observe that if the instance-level loss $\ell$ is convex, then $\atkloss$ is also convex\footnote{The pointwise maximum and the sum of convex functions preserve convexity.} and it can be efficiently optimized numerically. 
Convexity is indeed a property enjoyed by most loss functions such as SSE (for regression) and Log-Loss (for classification).
This allows one to overcome the exploration of the exponential number of attack configurations, still finding the optimal solution (up to numerical approximation).


Given the best predictions $\hat{y}_l,\hat{y}_r$, we can finally produce a binary split of $\dataset$ (as in Algorithm~\ref{alg:basictree}). To do this, we split those instances by applying the best adversarial moves, i.e., by assuming that every $\xy \in \rootunk$ is pushed into the left or right child so as to generate the largest loss. If the two children induce the same loss, then we assume the instance is not attacked.



\begin{definition}[Robust Splitting]\label{def:d-part}
Given a decision stump to be grown $\hat{t} = \node{f,v,\leaf{\hat{y}_l},\leaf{\hat{y}_r}}$ and an attacker $A$, the \emph{robust split} of the dataset $\dataset = \rootbinleft \cup \rootbinright$ is defined by:
\begin{itemize}
\item $\rootbinleft$ contains all the instances of $\rootleft$ and $\rootbinright$ contains all the instances of $\rootright$;
\item for each $\xy \in \rootunk$, the following rules apply:
\begin{itemize}
\item if $\ell(\hat{y}_l,y)>\ell(\hat{y}_r,y)$, then $\xy$ goes to $\rootbinleft$;
\item if $\ell(\hat{y}_l,y)<\ell(\hat{y}_r,y)$, then $\xy$ goes to $\rootbinright$;
\item if $\ell(\hat{y}_l,y)=\ell(\hat{y}_r,y)$, then $\xy$ goes to $\rootbinleft$ if $x_f\leq v$ and to $\rootbinright$ otherwise.
\end{itemize}
\end{itemize}
\end{definition}

\begin{example}[Robust Splitting]
Once identified $\hat{y}_l$ and $\hat{y}_r$ for the decision stump $\hat{t} = (p,1,\lambda(-1.5),\lambda(1.6))$ in Figure~\ref{fig:ex}.(c), the datasets obtained for the leaves by robust splitting are:
\begin{itemize}
    \item $\rootbinleft = \{(\vec{x0},-2), (\vec{x1},-1)\}$
    \item $\rootbinright = \{(\vec{x2},0), (\vec{x3},2), (\vec{x4},2), (\vec{x5},2), (\vec{x6},2)\}$
\end{itemize} 

Notice that, unlike a standard decision tree learning algorithm, the right partition contains the instance $\vec{x2}$ due to the presence of the attacker, even though such instance satisfies the root node test. 
\end{example}

To summarize, the ternary partitioning allows $\atkloss$ to be optimized for a given $(f,v)$ and dataset $\dataset$, so as to find the best tree-growing step by an exhaustive search over $f$ and $v$. Once this is done, the robust splitting allows the dataset $\dataset$ to be partitioned in order to feed the algorithm recursion on the left and right children of the newly created stump. Ultimately, the goal of robust splitting is to solve the min-max problem of Equation~\ref{eq:minmax} for a single tree-growing step, so as to find the best stump to be added to the tree, and push the attacked instances into the partition induced by the most harmful attack.

\subsection{Attack Invariance}
\label{sec:atk-invariance}

The optimization strategy described in Section~\ref{sec:robsplit} needs some additional refinement to provide a sound optimization of $\atkloss$ on the full dataset $\dataset$. When growing a new sub-tree at a leaf $\lambda$, we denote with $\dataset^\lambda$ the {\em local projection} of the full dataset at $\lambda$, i.e., the subset of the instances in $\dataset$ falling in $\lambda$ along the tree construction by applying the robust splitting strategy. The key observation now is that the robust splitting operates by assuming that the attacker behaves \emph{greedily}, i.e., by locally maximizing the generated loss, but as new nodes are added to the tree, new attack opportunities arise and different traversal paths towards different leaves may become more fruitful to the attacker. If this is the case, the robust splitting becomes unrepresentative of the possible attacker's moves and any learning decision made on the basis of such splitting turns out to be unsound, i.e., with no guarantee of minimizing $\atkloss$. Notice that this is a major design flaw of the algorithm proposed in~\cite{ChenZBH19}, and experimental evidence shows how the attacker can easily craft adversarial examples (see Section~\ref{sec:experiments}).

In the end, the computation of the best split for a given leaf $\lambda$ cannot be done just based on the local projection $\dataset^\lambda$, unless additional guarantees are provided. We thus enforce a security property called \emph{attack invariance}, which ensures that the tree construction steps preserve the correctness of the greedy assumptions made on the attacker's behavior. Given a decision tree $t$ and an instance $\xy \in \dataset$, we let $\Lambda^A(t,\xy)$ stand for the set of leaves of $t$ which are reachable by some attack $\vec{z} \in A(\vec{x})$ that generates the largest loss among $A(\vec{x})$.

Attack invariance requires that the tree construction steps preserves $\Lambda^A$,
in that the attacker has no advantage in changing the attack strategy which was optimal up to the previous step, thus recovering the soundness of the greedy construction. We define attack invariance during tree construction as follows.

\begin{definition}[Attack Invariance]
\label{def:a_inv}
Let $t$ be a decision tree and let $t'$ be the decision tree obtained by replacing a leaf $\lambda$ of $t$ with the new sub-tree $\node{f,v,\lambda_l,\lambda_r}$. We say that $t'$ satisfies \emph{attack invariance} for the dataset $\dataset$ and the attacker $A$ iff:
\[
\forall \xy \in \dataset^\lambda: \Lambda^A(t',\xy) \cap \{\lambda_l,\lambda_r\} \neq \emptyset.
\]
\end{definition}

The above definition states that, after growing a new sub-tree from $\lambda$, the set of the best options for the attacker must include the newly created leaves, so that the path originally leading to $\lambda$ still represents the most effective attack strategy against the decision tree.

\begin{example}[Attack Invariance]
\label{ex:attack invariance}
Let $t$ be the decision tree of Figure~\ref{fig:ex}.(c). Figure~\ref{fig:ex}.(d) shows an example where adding a new sub-tree to $t$ leads to a decision tree $t'$ which breaks the attack invariance property. Indeed, we have $\Lambda^A(t', (\vec{x2},0)) = \{\lambda(-1.5)\}$, which contains neither $\lambda(1)$, nor $\lambda(2)$. Notice that the best attack strategy has indeed changed with respect to $t$, as leaving $\vec{x2}$ unaltered now produces a larger loss (2.25) than the originally strongest attack (1.0). 
\end{example}

We enforce attack invariance by introducing a set of \emph{constraints} into the optimization problem  of Equation~\ref{eq:optimization}. Suppose that the new sub-tree $\node{f,v,\leaf{\hat{y}_l},\leaf{\hat{y}_r}}$ replaces the leaf $\lambda$ and that an instance $\xy \in \dataset^\lambda$ is placed in the right child by robust splitting, because one of its corruptions traverses the threshold $v$ and $\ell(\hat{y}_r,y)\ge\ell(\hat{y}_l,y)$. 
Then, attack invariance is granted if, whenever the leaves $\leaf{\hat{y}_l}$ and $\leaf{\hat{y}_r}$ are later replaced by sub-trees $t_l$ and $t_r$, there exists an attack $\vec{z} \in A(\vec{x})$ that falls into a leaf of $t_r$ generating a loss larger than (or equal to) the loss of any other attack falling in $t_l$. We enforce such constraint during the recursive tree building process as follows.
The requirement $\ell(\hat{y}_r,y)\ge\ell(\hat{y}_l,y)$ is transformed in 
the pair of constraints $\ell(t_r(\vec{x}),y)\ge\gamma$ and $\ell(t_l(\vec{x}),y)\le \gamma$,
where $\gamma=\min\{\ell(\hat{y}_r,y),\ell(\hat{y}_l,y)\}$. These two constraints
are propagated respectively into the recursion on the right and left children.
As long as any sub-tree $t_r$ replacing $\lambda(\hat{y}_r)$ satisfies the constraint 
$\ell(t_r(\vec{x}),y)\ge\gamma$ and any sub-tree $t_l$ replacing $\lambda(\hat{y}_l)$ satisfies the constraint $\ell(t_l(\vec{x}),y)\le\gamma$, the attacker has no advantage in changing the original attack strategy, hence attack invariance is enforced. 

To implement this mechanism, each leaf $\lambda$ is extended with a set of constraints, which is initially empty for the root of the tree. When $\lambda$ is then split upon tree growing, the constraints therein are included in the optimization problem of Equation~\ref{eq:optimization} to determine the best predictions $\hat{y}_l,\hat{y}_r$ for the new leaves. These constraints are then (partially) propagated to the new leaves and new constraints are generated for them based on the following definition, which formalizes the previous intuition.

\begin{definition}[Constraints Propagation and Generation]\label{def:c-part}
Let $\lambda$ be a leaf to be replaced with sub-tree $\hat{t} = \node{f,v,\leaf{\hat{y}_l},\leaf{\hat{y}_r}}$ and let $\constr$ be its set of constraints. The sets of constraints $\Cl$ and $\Cr$ for the two new leaves are defined by:\footnote{We use the symbol $\lessgtr$ to stand for either $\le$ or $\ge$ when the distinction is unimportant.}
\begin{itemize}
\item if $\ell(t(\vec{x}),y) \lessgtr \gamma \in \constr$ and there exists $\vec{z} \in A(\vec{x})$ such that $z_f\le v$, then $\ell(t_l(\vec{x}),y) \lessgtr \gamma$ is added to $\Cl$;

\item if $\ell(t(\vec{x}),y) \lessgtr \gamma \in \constr$ and there exists $\vec{z} \in A(\vec{x})$ such that $z_f> v$, then $\ell(t_r(\vec{x}),y) \lessgtr \gamma$ is added to $\Cr$;

\item if $\xy \in \atkunk \cap \binleft$, then $\ell(t_l(\vec{x}),y)\ge\loss(\hat{y}_r,y)$ is added to $\Cl$ and $\ell(t_r(\vec{x}),y)\le\loss(\hat{y}_r,y)$ is added to $\Cr$;

\item if $\xy \in \atkunk \cap \binright$, then $\ell(t_l(\vec{x}),y)\le\loss(\hat{y}_l,y)$ is added to $\Cl$ and $\ell(t_r(\vec{x}),y)\ge\loss(\hat{y}_l,y)$ is added to $\Cr$.
\end{itemize}
\end{definition}

\begin{example}[Enforcing Constraints] 
The tree in Fig.~\ref{fig:ex}.(e) is generated by enforcing a constraint on the loss of $\vec{x2}$. After splitting the root, the constraint $\ell(t_r(\vec{x2}),0) \ge \loss(\hat{y}_l,0)$ is generated for the right leaf of the tree in Fig.~\ref{fig:ex}.(c), where $\loss(\hat{y}_l,0) = (-1.5 - 0)^2 = 2.25$. The solution of the {\em constrained} optimization problem on the right child of the tree in Fig.~\ref{fig:ex}.(c) finally grows two new leaves, generating the tree in Fig.~\ref{fig:ex}.(e). The difference from the tree in Fig.~\ref{fig:ex}.(d) is that the prediction on the left leaf of the right child of the root has been enforced to satisfy the required constraint. For this tree, the attacker has no gain in changing attack strategy over the previous step of the tree construction, shown in Figure~\ref{fig:ex}.(c). 
\\
More formally, while for the tree $t$ in Fig.~\ref{fig:ex}.(c) we have $\Lambda^A(t, (\vec{x2},0)) = \{\lambda(1.6)\}$, after growing $t$ with suitable constraints we obtain the tree $t'$ in Fig.~\ref{fig:ex}.(e), where the leaf $\lambda(1.6)$ has been substituted with a decision stump with the two new leaves $\{\lambda(1.5), \lambda(2)\}$. 
This entails $\Lambda^A(t', (\vec{x2},0)) = \{\lambda(-1.5), \lambda(1.5)\}$, where $\Lambda^A(t', (\vec{x2},0)) \cap \{\lambda(1.5), \lambda(2)\} = \{\lambda(1.5)\} \neq \emptyset$, thus satisfying the attack invariance property of Definition~\ref{def:a_inv}.
\end{example}

Constraints grant attack invariance at the cost of reducing the space of the possible solutions for tree-growing. Nevertheless, in the experimental section we show that this property does not prevent the construction of robust decision trees that are also accurate in absence of attacks.

\subsection{Tree Learning Algorithm}
\label{sec:pseudocode}

Our \treant{} construction is summarized in Algorithm~\ref{alg:robusttree}. The core of the logic is in the call of the {\sc TSplit} function (line 3), which takes as input a dataset $\dataset$, an attacker $A$ and a set of constraints $\constr$ initially empty, and implements the construction detailed along the present section. The construction terminates when it is not possible to further reduce $\atkloss$ (line 4). 

Function {\sc TSplit} is summarized in Algorithm~\ref{alg:treantsplit}. Specifically, the function returns the sub-tree minimizing the loss under attack $\atkloss$ on $\dataset$ subject to the constraints $\constr$, based on the ternary partitioning (lines 2-3). It then splits $\dataset$ by means of the robust splitting strategy (lines 4-5) and returns new sets of constraints (lines 6-7), which are used to recursively build the left and right sub-trees. 
The optimization problem 
(line~2) is numerically solved via the \texttt{scipy} implementation of the SLSQP (Sequential Least SQuares Programming) method, which allows the minimization of a function subject to inequality constraints, like the constraint set $\constr$ generated/propagated by \treant\ during tree growing.  


\begin{algorithm}[t]
\caption{\treant}
\label{alg:robusttree}
\begin{algorithmic}[1]
\State {\bfseries Input:} {training data $\dataset$, attacker $A$, constraints $\constr$}

\State $\hat{y} \gets \argmin_y \atkloss(\leaf{y},\dataset)$ subject to $\constr$

\State $\node{f,v,\leaf{\hat{y}_l},\leaf{\hat{y}_r}},\dataset_l,\dataset_r,\constr_l,\constr_r \gets$ 
 {\sc TSplit}$(\dataset,A,\constr)$

\If{$\atkloss(\node{f,v,\leaf{\hat{y}_l},\leaf{\hat{y}_r}},\dataset) < \atkloss(\leaf{\hat{y}},\dataset)$}
	\State {$t_l \gets$ \treant$(\dataset_l,A,\constr_l)$}
	\State {$t_r \gets$ \treant$(\dataset_r,A,\constr_r)$}
    \State {\bf return} $\node{f,v,t_l,t_r}$
\Else
    \State {\bf return} $\leaf{\hat{y}}$
\EndIf
\end{algorithmic}
\end{algorithm}

\begin{algorithm*}[t]
\caption{\textsc{TSplit}} 
\label{alg:treantsplit}
\begin{algorithmic}[1]
\State {\bfseries Input:} training data $\dataset$, attacker $A$, constraints $\constr$
\Statex \Comment{Build a set of candidate tree nodes $\mathcal{N}$ using the ternary partitioning to optimize $\atkloss$}
\State{$\mathcal{N} \gets \{\ \sigma(f,v,\leaf{\hat{y}_l}, \leaf{\hat{y}_r}) ~|~ f \in [1,d] ~\wedge $
$\exists \xy \in \dataset: x_f = v ~\wedge $} \label{alg:search}

\Statex \hfill
$\begin{array}{rcl}
\\
\hat{y}_l,\hat{y}_r & = &\argmin\limits_{y_l, y_r}\ \sum\limits_{\xy \in \rootleft} \ell({y}_l,y)\ +
\sum\limits_{\xy \in \rootright} \ell({y}_r,y)\ + \sum\limits_{\xy \in \rootunk} \max \{\ell({y}_l,y), \ell({y}_r,y)\} \\ \\
 &  & \mathrm{subject}\ \mathrm{to}\ \constr 
\end{array}$
\Statex \qquad\ $\}$

\Statex\Comment{Select the candidate node $\hat{t} \in \mathcal{N}$ which minimizes the loss $\atkloss$ on the training data $\dataset$}
\State{$\hat{t} = \argmin_{t \in \mathcal{N}} \atkloss(t, \dataset) = \sigma(f,v,\leaf{\hat{y}_l},\leaf{\hat{y}_r})$}

\Statex\Comment{Robust Splitting (see Definition~\ref{def:d-part})} 

\State $\dataset_l \gets \rootbinleft$
\State $\dataset_r \gets \rootbinright$

\Statex\Comment{Constraint Propagation and Generation (see Definition~\ref{def:c-part})} 

\State $\constr_l \gets \Cl$
\State $\constr_r \gets \Cr$

\State {\bf return} {$\hat{t},\dataset_l, \dataset_r, \constr_l, \constr_r$}

\end{algorithmic}
\end{algorithm*}

There is an important point worth discussing about the implementation of the algorithm. As careful readers may have noticed, the {\sc TSplit} function splits each leaf $\lambda$ by relying on the set of attacks $A(\vec{x})$ for all instances $\xy \in \dataset^\lambda$. Though one could theoretically pre-compute all the possible attacks against the instances in $\dataset$, this implementation would be very inefficient both in time and space, given the potentially huge number of instances and attacks. Our implementation, instead, \emph{incrementally} computes a {\em sufficient} subset of $A(\vec{x})$ along the tree construction. 

First, each instance $\xy$ is enriched with a cost annotation $k$, denoted by $\xyk$, initially set to 0 on the root. Such annotation keeps track of the cost of the adversarial manipulations performed to push $\xy$ into $\lambda$ during the tree construction. When splitting the leaf $\lambda$ on $(f,v)$, the algorithm generates only the attacks against the feature $f$ which enforce \emph{maximal} perturbations of $x_f$, as such maximal perturbations maximize the chance of crossing the threshold $v$ without incurring in any extra cost. Moreover, the attack generation assumes that $k$ was already spent from the attacker's budget to further reduce the number of possible attacks. When the instance $\xyk$ is pushed into the left or right partition of $\dataset^\lambda$ by robust splitting, the label $k$ is updated to $k+k'$, where $k'$ is the \emph{minimum cost} the attacker must spend to achieve the desired node outcome. The same idea is applied when propagating constraints, which are also associated with specific instances $\xy$ for which the computation of $A(\vec{x})$ is required.

Observe that this implementation assumes that only the cost of adversarial manipulations is relevant, not their magnitude, which is still sound when none of the corrupted features is tested multiple times on the same path of the tree. We enforce such restriction during the tree construction, which further regularizes the growing of the tree. Since we are eventually interested in decision tree ensembles, this does not impact on the performance of whole trained models.

\section{Experimental Evaluation}
\label{sec:experiments}
\subsection{Methodology}
\label{subsec:methodology}
We compare the performance of classifiers trained by different learning algorithms: two standard approaches, i.e., Random Forest~\cite{RandomF01} (RF) and Gradient Boosting Decision Trees~\cite{friedman2001greedy} (GBDT) as provided by the LightGBM\footnote{\url{https://github.com/microsoft/LightGBM}} framework; two state-of-the-art adversarial learning techniques, i.e., Adversarial Boosting~\cite{KantchelianTJ16} (AB) and Robust Trees~\cite{ChenZBH19} (RT); and a Random Forest of trees trained using the proposed \treant{} algorithm (RF-\treant).\footnote{The source code of \treant{} is available at \url{https://github.com/omitted-for-anonymous-review}} Notice that the original implementation of AB exploited a heuristic algorithm to find good adversarial examples, which does not guarantee to find the most damaging attack. Our own implementation of AB, which is built on top of LightGBM, exploits the white-box attack generation method described in Section~\ref{sec:attacks} to find the {\em best} adversarial examples. In this regard, our implementation is thus more effective than the original algorithm.

We perform our experimental evaluation on three publicly available datasets, using three standard validity measures: accuracy, macro F1 and ROC AUC. We compute all measures both in absence of attacks and under attack, using our white-box attack generation method. \cla{We used a 60-20-20 train-validation-test split through stratified sampling. 
Hyper-parameter tuning on the validation data
was conducted to set the number of trees ($\le 100$), number of leaves ($\{8, 32, 256\}$) and learning rate ($\{0.01, 0.05, 0.1\}$) of the various ensembles so as to maximize ROC AUC.
All the results reported below were measured on the test data.}
Observe that all the compared adversarial learning techniques are parametric with respect to the budget granted to the attacker, modeling his power: we consider multiple instances of such budget both for training (\emph{train budget}) and for testing (\emph{test budget}).

The goal of our experimental evaluation is answering three key questions:
\begin{enumerate}
\item What is the performance of standard learning approaches like RF and GBDT when they are adopted in an adversarial setting?
\item What is the performance improvement achieved by the adoption of adversarial learning techniques for different test budgets?
\item What is the importance of the training budget on the performance of adversarial learning techniques?
\end{enumerate}

\subsection{Datasets and Threat Models}
\label{subsec:datasets}
We perform our experimental evaluation on three datasets: $(i)$ Census Income,\footnote{\url{https://archive.ics.uci.edu/ml/datasets/census+income}} $(ii)$ Wine Quality,\footnote{\url{https://www.kaggle.com/c/uci-wine-quality-dataset/data}} and $(iii)$ Default of Credit Cards.\footnote{\url{https://archive.ics.uci.edu/ml/datasets/default+of+credit+card+clients}} In the following, we refer to such datasets as \census, \wine, and \credit, respectively. Their main statistics are shown in Table~\ref{tab:datasets}; notice that each dataset is associated with a binary classification task.\footnote{The \wine\ dataset was originally conceived for a multiclass classification problem; we turned that into a binary one, where the positive class identifies good-quality wines (i.e., those whose quality is at least 6, on a 0-10 scale) and the negative class contains the remaining instances.}

\begin{table}[t]
\centering
\renewcommand{\arraystretch}{1.2}
\caption{Main statistics of the datasets used in our experiments.}
\label{tab:datasets}
\begin{tabular}{c|c|c|c|}
\cline{2-4}
 & \census & \wine  & \credit \\ \hline
 \multicolumn{1}{|c|}{n. of instances} & 45,222 &6,497  &30,000  \\ \hline
\multicolumn{1}{|c|}{n. of features} & 13  &12  &24  \\ \hline
 \multicolumn{1}{|c|}{class distribution (pos.$\div$neg. \%)} & 25$\div$75  & 63$\div$37 & 22$\div$78 \\ \hline
\end{tabular}
\end{table}

We therefore design three different threat models by means of a set of rewriting rules indicating the attacker capabilities, with each set tailored to a given dataset. The features targeted by those rules have been selected after a preliminary data exploration stage, where we investigated the importance and data distribution of all the features.

In the case of \census, we define six rewriting rules: $(i)$ if a citizen never worked, he can pretend that he actually works without pay; $(ii)$ if a citizen is divorced or separated, he can pretend that he never got married; $(iii)$ a citizen can present his occupation as a generic ``other service''; $(iv)$ a citizen can cheat on his education level by lowering it by 1; $(v)$ a citizen can add up to \$2,000 to his capital gain; $(vi)$ a citizen can add up to 4 hrs per week to his working hours. We let $(i)$,$(ii)$, and $(iii)$ cost 1, $(iv)$ cost 20, $(v)$ cost 50, and finally $(vi)$ cost 100 budget units. We consider 30, 60, 90, and 120 as possible values for the budget.

In the case of \wine, we specify four rewriting rules: $(i)$ the alcohol level can be increased by 0.5\% if its original value is less than  11\%; $(ii)$ the residual sugar can be decreased by 0.25 g/L if it is already greater than or 2 g/L; $(iii)$ the volatile acidity can be reduced by 0.1 g/L if it is already greater than 0.25 g/L; $(iv)$ free sulfur dioxide reduced by -2 g/L if it is already greater than 25 g/L. We let $(i)$ cost 20, $(ii)$ and $(iii)$ cost 30, and $(iv)$ cost 50 budget units. We consider 20, 40, 60, 80, 100, and 120 as possible values for the budget.

For \credit, the attacker is represented by three rewriting rules: $(i)$ the repayment status on August or September can be reduced by 1 month if the payment is delayed up to 5 months; $(ii)$ the amount of bill statement in September can be decreased by 4,000 NT dollars if it is between 20,000 and 500,000; and $(iii)$ the amount of given credit can be increased by 20,000 NT dollars if it is below 200,000. For each rule, a cost of 10 budget units is required. We consider 10, 30, 40, and 60 as possible budget values.

\begin{table*}[t]
\centering
\renewcommand{\arraystretch}{1.2}
\renewcommand\tabcolsep{5pt}
\caption{Comparison of adversarial learning techniques trained and attacked under the same budget. The table also shows the performance difference between RF-\treant{} and the best competitor.}
\label{tab:train-test-budget}
\begin{tabular}{ccc|ccc|ccc|cccccc|}
\cline{4-15}
& & & \multicolumn{3}{ c|| }{AB} & \multicolumn{3}{ c|| }{RT} & \multicolumn{6}{ c| }{RF-\treant} \\ 
\cline{4-15}
& & & \multicolumn{1}{ c| }{Accuracy} & \multicolumn{1}{ c| }{$F_1$} & \multicolumn{1}{ c||}{ROC AUC} & \multicolumn{1}{ c| }{Accuracy} & \multicolumn{1}{ c| }{$F_1$} & \multicolumn{1}{ c||}{ROC AUC} &
\multicolumn{2}{ c| }{Accuracy} & \multicolumn{2}{ c| }{$F_1$} & \multicolumn{2}{ c|}{ROC AUC} \\ 
\cline{1-15}
\multicolumn{1}{ |c  }{\multirow{4}{*}{\census} } & 
\multicolumn{1}{ |c  }{\multirow{4}{*}{\rotatebox[origin=c]{90}{Budget}}} &
\multicolumn{1}{|c|}{30} & 
\multicolumn{1}{c|}{{\bf0.850}} & \multicolumn{1}{c|}{{\bf 0.783}} & \multicolumn{1}{c||}{{\bf 0.902}} & \multicolumn{1}{c|}{0.813} & \multicolumn{1}{c|}{0.692} & \multicolumn{1}{c||}{0.883} & \multicolumn{1}{c|}{{\bf0.850}} & \multicolumn{1}{c|}{+0.0\%} & \multicolumn{1}{c|}{0.773} & \multicolumn{1}{c|}{-1.3\%} & \multicolumn{1}{c|}{0.897} & \multicolumn{1}{c|}{-0.6\%} \\
\cline{3-15}
\multicolumn{1}{|c}{} & \multicolumn{1}{|c}{} & 
\multicolumn{1}{|c|}{60} & 
\multicolumn{1}{c|}{0.783} & \multicolumn{1}{c|}{0.690} & \multicolumn{1}{c||}{0.827} & \multicolumn{1}{c|}{0.810} & \multicolumn{1}{c|}{0.698} & \multicolumn{1}{c||}{0.871} & \multicolumn{1}{c|}{{\bf 0.845}} & \multicolumn{1}{c|}{+4.3\%} & \multicolumn{1}{c|}{{\bf 0.766}} & \multicolumn{1}{c|}{+9.7\%} & \multicolumn{1}{c|}{{\bf 0.894}} & \multicolumn{1}{c|}{+2.6\%} \\
\cline{3-15}
\multicolumn{1}{|c}{} & \multicolumn{1}{|c}{} & 
\multicolumn{1}{|c|}{90} & 
\multicolumn{1}{c|}{0.798} & \multicolumn{1}{c|}{0.705} & \multicolumn{1}{c||}{0.825} & \multicolumn{1}{c|}{0.775} & \multicolumn{1}{c|}{0.607} & \multicolumn{1}{c||}{0.855} & \multicolumn{1}{c|}{{\bf 0.845}} & \multicolumn{1}{c|}{+5.9\%} & \multicolumn{1}{c|}{{\bf 0.769}} & \multicolumn{1}{c|}{+9.1\%} & \multicolumn{1}{c|}{{\bf 0.893}} & \multicolumn{1}{c|}{+4.4\%} \\
\cline{3-15}
\multicolumn{1}{|c}{} & \multicolumn{1}{|c}{} & 
\multicolumn{1}{|c|}{120} & 
\multicolumn{1}{c|}{0.788} & \multicolumn{1}{c|}{0.694} & \multicolumn{1}{c||}{0.793} & \multicolumn{1}{c|}{0.744} & \multicolumn{1}{c|}{0.558} & \multicolumn{1}{c||}{0.528} & \multicolumn{1}{c|}{{\bf 0.842}}& \multicolumn{1}{c|}{+6.9\%} & \multicolumn{1}{c|}{{\bf 0.762}} & \multicolumn{1}{c|}{+9.8\%} & \multicolumn{1}{c|}{{\bf 0.887}} & \multicolumn{1}{c|}{+11.9\%} \\
\cline{1-15}
\multicolumn{1}{ |c  }{\multirow{6}{*}{\wine} } & 
\multicolumn{1}{ |c  }{\multirow{6}{*}{\rotatebox[origin=c]{90}{Budget}}} &
\multicolumn{1}{|c|}{20} & 
\multicolumn{1}{c|}{0.762} & \multicolumn{1}{c|}{0.737} & \multicolumn{1}{c||}{{\bf 0.824}} & \multicolumn{1}{c|}{0.734} & \multicolumn{1}{c|}{0.703} & \multicolumn{1}{c||}{0.795} & \multicolumn{1}{c|}{{\bf 0.764}} & \multicolumn{1}{c|}{+0.3\%} & \multicolumn{1}{c|}{{\bf 0.739}} & \multicolumn{1}{c|}{+0.3\%} & \multicolumn{1}{c|}{0.821}& \multicolumn{1}{c|}{-0.4\%} \\
\cline{3-15}
\multicolumn{1}{|c}{} & \multicolumn{1}{|c}{} & 
\multicolumn{1}{|c|}{40} & \multicolumn{1}{c|}{0.723} &
\multicolumn{1}{c|}{{\bf 0.689}} &
\multicolumn{1}{c||}{0.788} & \multicolumn{1}{c|}{0.623} & \multicolumn{1}{c|}{0.548} & \multicolumn{1}{c||}{0.662} & \multicolumn{1}{c|}{{\bf 0.728}} & \multicolumn{1}{c|}{+0.7\%} & \multicolumn{1}{c|}{{\bf 0.689}} & \multicolumn{1}{c|}{+0.0\%} & \multicolumn{1}{c|}{{\bf 0.802}}& \multicolumn{1}{c|}{+1.8\%} \\
\cline{3-15}
\multicolumn{1}{|c}{} & \multicolumn{1}{|c}{} & 
\multicolumn{1}{|c|}{60} & 
\multicolumn{1}{c|}{0.718} & \multicolumn{1}{c|}{{\bf 0.687}} & \multicolumn{1}{c||}{0.788} & \multicolumn{1}{c|}{0.552} & \multicolumn{1}{c|}{0.418} & \multicolumn{1}{c||}{0.522} & \multicolumn{1}{c|}{{\bf 0.720}} & \multicolumn{1}{c|}{+0.3\%} & \multicolumn{1}{c|}{0.680} & \multicolumn{1}{c|}{-1.0\%} & \multicolumn{1}{c|}{{\bf 0.798}}& \multicolumn{1}{c|}{+1.3\%} \\
\cline{3-15}
\multicolumn{1}{|c}{} & \multicolumn{1}{|c}{} & 
\multicolumn{1}{|c|}{80} & 
\multicolumn{1}{c|}{0.715} & \multicolumn{1}{c|}{0.680} & \multicolumn{1}{c||}{0.773} & \multicolumn{1}{c|}{0.566} & \multicolumn{1}{c|}{0.443} & \multicolumn{1}{c||}{0.561} & \multicolumn{1}{c|}{{\bf 0.728}} & \multicolumn{1}{c|}{+1.8\%} & \multicolumn{1}{c|}{{\bf 0.688}} & \multicolumn{1}{c|}{+1.2\%} & \multicolumn{1}{c|}{{\bf 0.800}} & \multicolumn{1}{c|}{+3.5\%} \\
\cline{3-15}
\multicolumn{1}{|c}{} & \multicolumn{1}{|c}{} & 
\multicolumn{1}{|c|}{100} & 
\multicolumn{1}{c|}{0.702} & \multicolumn{1}{c|}{0.668} & \multicolumn{1}{c||}{0.761} & \multicolumn{1}{c|}{0.559} & \multicolumn{1}{c|}{0.429} & \multicolumn{1}{c||}{0.553} & \multicolumn{1}{c|}{{\bf 0.727}} & \multicolumn{1}{c|}{+3.6\%} & \multicolumn{1}{c|}{{\bf 0.687}} & \multicolumn{1}{c|}{+2.8\%} & \multicolumn{1}{c|}{{\bf 0.796}} & \multicolumn{1}{c|}{+4.6\%} \\
\cline{3-15}
\multicolumn{1}{|c}{} & \multicolumn{1}{|c}{} & 
\multicolumn{1}{|c|}{120} & 
\multicolumn{1}{c|}{0.677} & \multicolumn{1}{c|}{0.636} & \multicolumn{1}{c||}{0.732} & \multicolumn{1}{c|}{0.568} & \multicolumn{1}{c|}{0.431} & \multicolumn{1}{c||}{0.544} & \multicolumn{1}{c|}{{\bf 0.728}} & \multicolumn{1}{c|}{+7.5\%} & \multicolumn{1}{c|}{{\bf 0.688}} & \multicolumn{1}{c|}{+8.2\%} & \multicolumn{1}{c|}{{\bf 0.801}} & \multicolumn{1}{c|}{+9.4\%} \\
\cline{1-15}
\multicolumn{1}{ |c  }{\multirow{4}{*}{\credit} } & 
\multicolumn{1}{ |c  }{\multirow{4}{*}{\rotatebox[origin=c]{90}{Budget}}} &
\multicolumn{1}{|c|}{10} & 
\multicolumn{1}{c|}{0.811} & \multicolumn{1}{c|}{0.644} & \multicolumn{1}{c||}{0.749} & \multicolumn{1}{c|}{0.799} & \multicolumn{1}{c|}{0.610} & \multicolumn{1}{c||}{0.748} & \multicolumn{1}{c|}{{\bf 0.816}} & \multicolumn{1}{c|}{+0.6\%} & \multicolumn{1}{c|}{{\bf 0.656}} & \multicolumn{1}{c|}{+1.9\%} & \multicolumn{1}{c|}{{\bf 0.765}} & \multicolumn{1}{c|}{+2.1\%} \\
\cline{3-15}
\multicolumn{1}{|c}{} & \multicolumn{1}{|c}{} & 
\multicolumn{1}{|c|}{30} & 
\multicolumn{1}{c|}{0.786} & \multicolumn{1}{c|}{0.544} & \multicolumn{1}{c||}{0.661} & \multicolumn{1}{c|}{0.763} & \multicolumn{1}{c|}{0.457} & \multicolumn{1}{c||}{0.655} & \multicolumn{1}{c|}{{\bf 0.810}} & \multicolumn{1}{c|}{+3.1\%} & \multicolumn{1}{c|}{{\bf 0.617}} & \multicolumn{1}{c|}{+13.4\%} & \multicolumn{1}{c|}{{\bf 0.745}} & \multicolumn{1}{c|}{+12.7\%} \\
\cline{3-15}
\multicolumn{1}{|c}{} & \multicolumn{1}{|c}{} &
\multicolumn{1}{|c|}{40} & 
\multicolumn{1}{c|}{0.784} & \multicolumn{1}{c|}{0.554} & \multicolumn{1}{c||}{0.660} & \multicolumn{1}{c|}{0.759} & \multicolumn{1}{c|}{0.438} & \multicolumn{1}{c||}{0.632} & \multicolumn{1}{c|}{{\bf 0.808}} & \multicolumn{1}{c|}{+3.1\%} & \multicolumn{1}{c|}{{\bf 0.618}} & \multicolumn{1}{c|}{+11.6\%} & \multicolumn{1}{c|}{{\bf 0.744}} & \multicolumn{1}{c|}{+12.7\%} \\
\cline{3-15}
\multicolumn{1}{|c}{} & \multicolumn{1}{|c}{} & 
\multicolumn{1}{|c|}{60} & 
\multicolumn{1}{c|}{0.777} & \multicolumn{1}{c|}{0.533} & \multicolumn{1}{c||}{0.622} & \multicolumn{1}{c|}{0.759} & \multicolumn{1}{c|}{0.436} & \multicolumn{1}{c||}{0.613} & \multicolumn{1}{c|}{{\bf 0.809}} & \multicolumn{1}{c|}{+4.1\%} & \multicolumn{1}{c|}{{\bf 0.616}} & \multicolumn{1}{c|}{+15.6\%} & \multicolumn{1}{c|}{{\bf 0.744}} & \multicolumn{1}{c|}{+19.6\%} \\
\cline{1-15}
\end{tabular}
\end{table*}

\subsection{Experimental Results}
\cla{We discuss below the three questions stated in Section~\ref{subsec:methodology}.}
 
\subsubsection{Attacking standard decision tree ensembles}
\begin{figure}[tb]
    \centering
    \includegraphics[width=.5\textwidth]{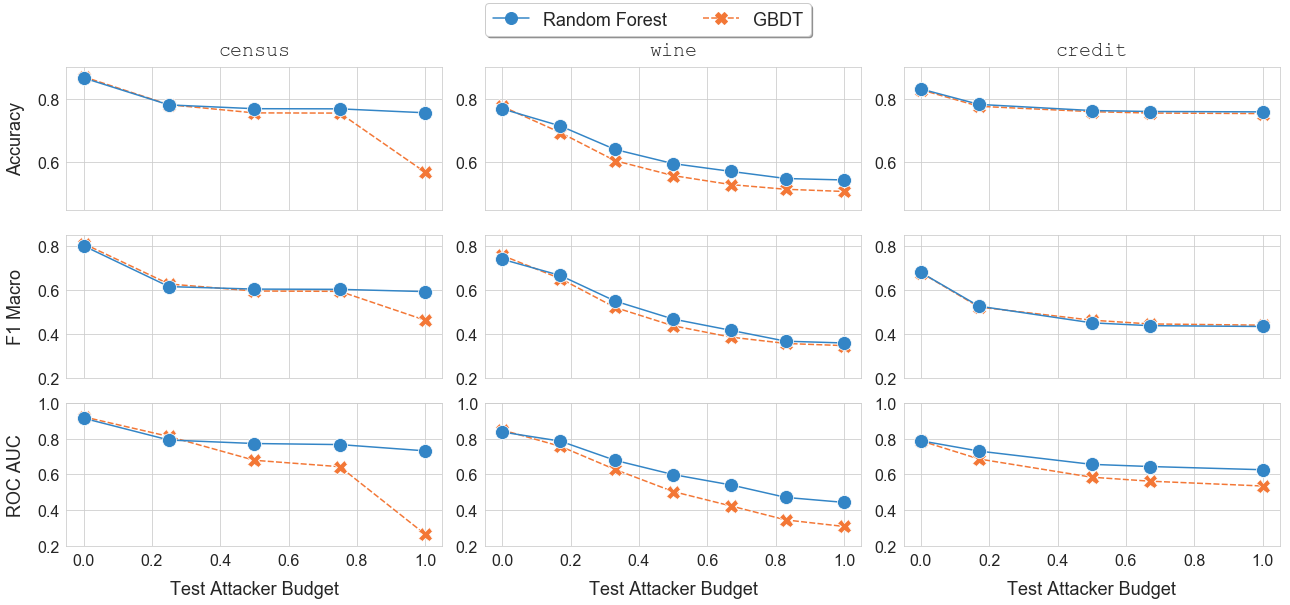}
    \caption{The impact of the attacker on RF and GBDT.}
    \label{fig:gbdt-vs-rf}
\end{figure}

In Figure~\ref{fig:gbdt-vs-rf}, we show how the accuracy, F1, and ROC AUC of standard ensembles of decision trees trained by RF and GBDT change in presence of attacks. The $x$-axis indicates the testing budget of the attacker, normalized in the range $[0,1]$, with a value of $0$ denoting the unattacked scenario. 

Two main findings appear from the plots. First, both GBDT and RF are severely impacted when they are attacked, and their performance deteriorates to the point of turning them into almost random classifiers already when the attacker spends just half of the maximum budget, e.g., in the case of the \wine\ dataset. On that dataset, the drop of ROC AUC ranges from -25.8\% to -40.6\% for GBDT and from -15.5\% to -28.4\% for RF, when the attacker is supplied just half of the budget. Second, RF typically behaves better than GBDT on all the validity measures, with a few cases where the improvement is very significant. A possible explanation of this phenomenon is that RF usually exhibits better generalization performance, while GBDT is known to be more susceptible to jiggling data, therefore more likely to overfit~\cite{nawar2017sensors}. Since robustness to adversarial examples in a way resembles the ability of a model to generalize, RF is less affected by the attacker than GBDT. Still, the performance drop under attack is so massive even for RF that none of the traditional methods can be reliably adopted in an adversarial setting.

The higher resiliency of RF to adversarial examples motivated our choice to deploy \treant{} on top of such ensemble method in our implementation. It is worth remarking though that \treant{} is still general enough to be plugged into other frameworks for ensemble tree learning.

\subsubsection{Robustness of adversarial learning techniques}
We now measure the benefit of using adversarial learning techniques to contrast the impact of evasion attacks at test time. More specifically, we validate the robustness of our method in comparison with the two state-of-the-art adversarial learning methods Adversarial Boosting (AB) and Robust Trees (RT).
\cla{Note that the authors of \cite{ChenZBH19} did not experimentally compare RT against AB in their original work.}

We first investigate how robust a model is when it is targeted by an attacker with a test budget exactly matching the training budget. This simulates the desirable scenario where the threat model was defined accurately, i.e., each model is trained knowing the actual attacker capabilities. Table~\ref{tab:train-test-budget} shows the results obtained by the different adversarial learning techniques for the different training/test budgets. It is clear how our method outperforms its competitors, basically for all measures and datasets. Most importantly, the superiority of our approach becomes even more pronounced as the strength of the attacker grows. For example, the percentage improvement in ROC AUC over AB on the \credit{} dataset amounts to 2.1\% for budget 10, while this improvement grows to 19.6\% for budget 60. It is also worth noticing that the performance of RT is consistently worse than that of AB.

\begin{figure*}[t]
    \centering
    \includegraphics[width=\textwidth]{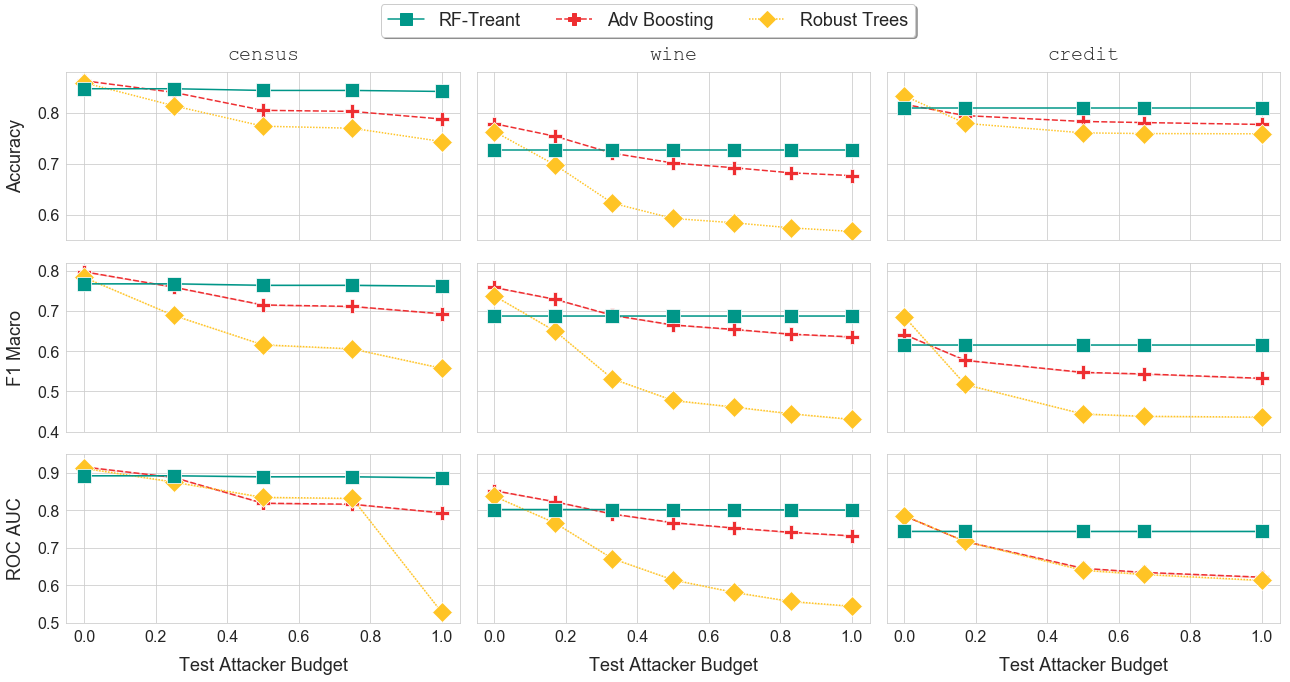}
    \caption{Comparison of adversarial learning techniques for different test budgets and maximum train budget.}
    \label{fig:fixed-train}
\end{figure*}

The second analysis we carry out considers the case of adversarial learning techniques trained with the maximum available budget. We use security evaluation curves to measure how the performance of the compared methods changes when the test budget given to the attacker increases up to the maximum available.
The results are shown in Figure~\ref{fig:fixed-train}, where we normalized the test budget in the range $[0,1]$.

\begin{figure*}[t!]
    \centering
    \includegraphics[width=\textwidth]{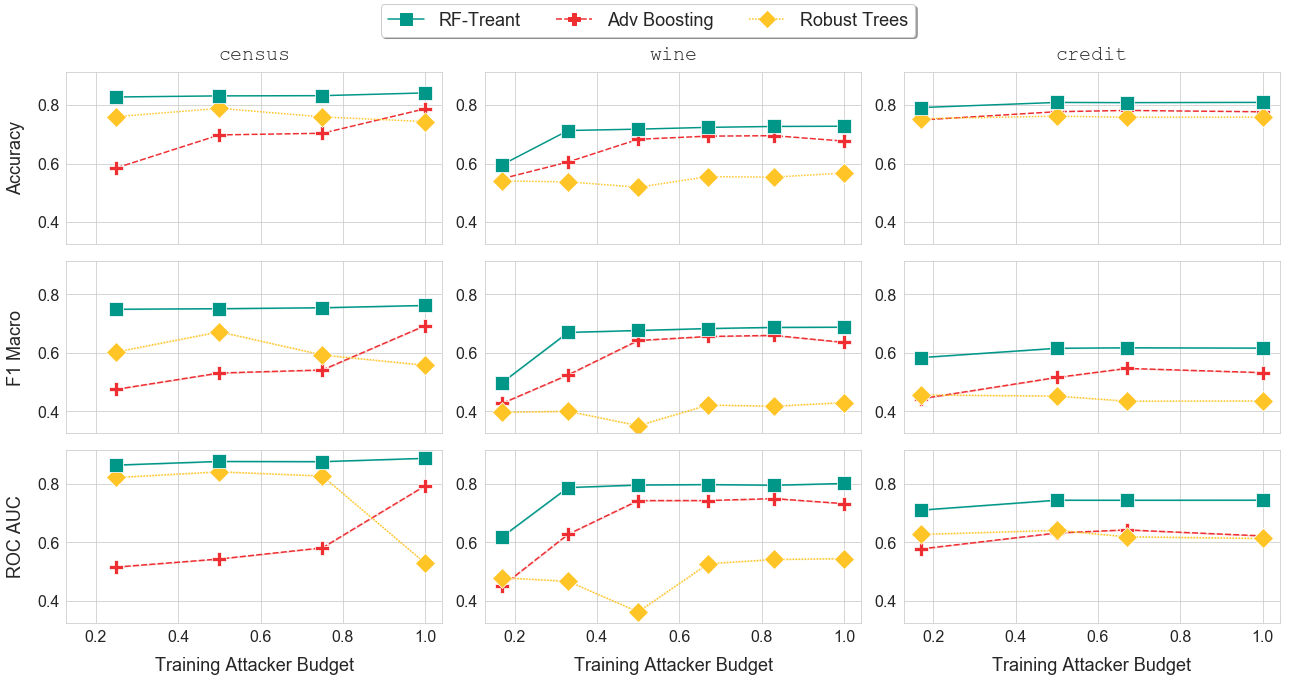}
    \caption{Comparison of adversarial learning techniques for  different train budgets and  maximum test budget.}
\label{fig:fixed-test}
\end{figure*}

Two main comments can be made from the plots. First, our method constantly outperforms its competitors on all datasets and measures, especially when the attacker gets stronger. The price to pay for this increased protection is just a slight performance degradation in the unattacked setting, which is largely compensated under attack. Indeed, the performance of our method is nearly constant and insensitive to variations in the attacker's budget, which is extremely useful when such information is hard to quantify exactly. Second, we observe that AB is usually more robust than RT. We believe that RT suffers from its heuristic splitting strategy, which is not smart enough to counteract the full spectrum of possible attacks, and the lack of attack invariance. There are indeed a few cases where the performance of RT upon attack is comparable to the performance of traditional GBDT.

\subsubsection{Impact of training budget}
A last intriguing aspect to consider is how much adversarial learning techniques are affected by the assumptions made on the attacker's capabilities upon learning, i.e., the training budget.
Figure~\ref{fig:fixed-test} is essentially the dual of Figure~\ref{fig:fixed-train}, where we consider the strongest possible attacker (with the largest test budget) and we analyze how much models learned with different (smaller or equal) training budgets  are able to respond to evasion attempts.

We draw the following observations. First, our method leads to the most robust models for all measures and datasets, irrespective of the budget used for training. Moreover, our method is the one which most evidently presents a healthy, expected trend: the greater the training budget used to learn the model, the better its performance under attack. This trend eventually reaches its peak when the training budget matches the test budget. Also AB shows a similar trend, yet it suffers from a slow start before reaching its best performance, which is however worse than our method. RT is the method which shows the most unpredictable behavior, as its performance fluctuates up and down, and sometimes suddenly drops. This is likely due to the fact that the heuristic it implements is too shortsighted with respect to the set of all the attacks and the lack of attack invariance. Finally, we remark a last appealing, distinctive aspect of our method: even when the training uses a significantly smaller budget than the one used by the attacker at test time, it already achieves nearly optimal performances. The same is not true for its competitors, which complicates their deployment in real-world settings.

\section{Conclusion}

This paper proposes \treant, a new adversarial learning algorithm that is able to grow decision trees that are resilient against evasion attacks. \treant{} is the first algorithm which greedily, yet soundly, minimizes an evasion-aware loss function, which captures the attacker's goal of maximizing prediction errors. Our experiments, conducted on three publicly available datasets, confirm that \treant{} produces accurate tree ensembles, which are extremely robust against evasion attacks. Compared to the state of the art,
\treant{} exhibits a ROC AUC improvement against the strongest attacker ranging from $\approx 10\%$ to $\approx 20\%$.

As future work, we plan to revise our decision tree construction to make it aware of its deployment inside an ensemble; in other words, we aim at exploiting the information that the currently grown ensemble is particularly strong or weak against some classes of attacks to guide the construction of the next member of the ensemble. We also plan to evaluate our learning technique against regression datasets to get an additional quantitative evaluation of its security benefits. Finally, we want to investigate the combined use of standard decision trees and decision trees trained using \treant{} in the same ensemble, to achieve the optimal trade-off between accuracy in the unattacked setting and resilience to attacks.

\bibliographystyle{IEEEtran}
\bibliography{main}

\end{document}